\documentclass[10pt,journal,compsoc]{IEEEtran-customize}

\usepackage{cite}
\usepackage{amsmath}
\usepackage{amssymb}
\usepackage{xcolor}

\usepackage{graphicx}
\usepackage{booktabs}
\usepackage{multirow}
\usepackage{makecell}
\usepackage{multicol}
\usepackage{adjustbox}
\usepackage{hhline}

\usepackage{enumitem}
\usepackage{url}

\usepackage{subcaption}

\newcolumntype{L}[1]{>{\raggedright\arraybackslash}p{#1}}
\newcolumntype{C}[1]{>{\centering\arraybackslash}p{#1}}
\newcolumntype{R}[1]{>{\raggedleft\arraybackslash}p{#1}}

\definecolor{purple}{RGB}{128, 0, 128}

\definecolor{purple}{RGB}{128, 0, 128}
\definecolor{LightRed}{rgb}{1,0.92,0.92}
\definecolor{LightOrange}{rgb}{1,0.95,0.88}
\definecolor{LightYellow}{rgb}{1.0,1.0,0.84}
\definecolor{LightGreen}{rgb}{0.9,1.0,0.88}
\definecolor{LightCyan}{rgb}{0.9,1,1}
\definecolor{LightBlue}{rgb}{0.9,0.94,1}
\definecolor{LightIndigo}{rgb}{0.92,0.9,1}
\definecolor{LightMagenta}{rgb}{0.96,0.86,1}
\definecolor{DirtyWhite}{rgb}{0.96,0.96,0.96}

\DeclareSymbolFont{extraup}{U}{zavm}{m}{n}
\DeclareMathSymbol{\varheart}{\mathalpha}{extraup}{86}
\DeclareMathSymbol{\vardiamond}{\mathalpha}{extraup}{87}
\DeclareMathSymbol{\varclubsuit}{\mathalpha}{extraup}{88}

\begin{document}

\title{Frustratingly Easy Task-aware Pruning for\\Large Language Models}

\author{
    Yuanhe Tian$^{\varheart}$, \hspace{0.1cm}
    Junjie Liu$^{{\spadesuit}}$, \hspace{0.1cm}
    Xican Yang$^{{\spadesuit}}$, \hspace{0.1cm}
    Haishan Ye$^{{\Diamond}}$, \hspace{0.1cm}
    Yan Song$^{{\spadesuit}*}$ \\
    \vspace{0.2cm}
    $^{\varheart}$Zhongguancun Institute of Artificial Intelligence \hspace{0.1cm}
    $^{\spadesuit}$University of Science and Technology of China \\
    $^{{\Diamond}}$Xi’an Jiaotong University\\
    \vspace{0.1cm}
    $^{\varheart}$\texttt{tianyuanhe@zgci.ac.cn} \hspace{0.1cm}
    $^{\spadesuit}$\texttt{\{ljj19937347730, yxc15759879600\}@mail.ustc.edu.cn} \\
    $^{{\Diamond}}$\texttt{yehaishan@xjtu.edu.cn}
    \hspace{0.1cm}
    $^{\spadesuit}$\texttt{clksong@gmail.com}
}

\IEEEtitleabstractindextext{%

\begin{abstract}
Pruning provides a practical solution to reduce the resources required to run large language models (LLMs) to benefit from their effective
capabilities as well as control their cost for training and inference.
Research on LLM pruning often ranks the importance of LLM parameters using their magnitudes and calibration-data activations and removes (or masks) the less important ones, accordingly reducing LLMs' size.
However, these approaches primarily focus on preserving the LLM’s ability to generate fluent sentences, while neglecting performance on specific domains and tasks.
In this paper, we propose a simple yet effective pruning approach for LLMs that preserves task-specific capabilities while shrinking their parameter space.
\textcolor{black}{
We first analyze how conventional pruning minimizes loss perturbation under general-domain calibration and extend this formulation by incorporating task-specific feature distributions into the importance computation of existing pruning algorithms.
Thus, our framework computes separate importance scores using both general and task-specific calibration data, partitions parameters into shared and exclusive groups based on activation-norm differences, and then fuses their scores to guide the pruning process.
This design enables our method to integrate seamlessly with various foundation pruning techniques and preserve the LLM’s specialized abilities under compression.
}
Experiments on widely used benchmarks demonstrate that our approach is effective and consistently outperforms the baselines with identical pruning ratios and different settings.
\end{abstract}

}

\maketitle
\IEEEdisplaynontitleabstractindextext
\IEEEpeerreviewmaketitle

\makeatletter
\def\@IEEEcompsocmakefnmark{\hbox{\normalfont\@thefnmark\ }}
\long\def\@makefntext#1{\parindent 1em\indent\hbox{\@IEEEcompsocmakefnmark}#1}
\makeatother

\makeatletter
\def\@IEEEcompsocmakefnmark{\hbox{\normalfont\@thefnmark.\ }}
\long\def\@makefntext#1{\parindent 1em\indent\hbox{\@IEEEcompsocmakefnmark}#1}
\makeatother

\renewcommand{\thefootnote}{\arabic{footnote}}

\renewcommand{\thefootnote}{\fnsymbol{footnote}}
% \footnotetext[1]{Equal contributions.}
\footnotetext[1]{Corresponding Author.}
\renewcommand{\thefootnote}{\arabic{footnote}}

\section{Introduction}

Large language models (LLMs) present outstanding performance across a diverse set of artificial intelligence (AI) tasks \cite{brown2020language,chekroud2021promise,wei2022emergent,achiam2023gpt,taori2023alpaca,bubeck2023sparks,touvron2023llama-2,hadi2023large,achiam2023gpt,tian-etal-2024-chimed}.
However, their substantial size and intensive computational requirements pose significant challenges for deployment in resource-constrained environments \cite{wu2016quantized,zhou2019edge,strubell2020energy,hoffmann2022training,ainslie2023gqa}.
Pruning these models, i.e., selectively removing redundant or low-impact parameters, has emerged as a key strategy to lower inference latency and memory usage while preserving model capabilities \cite{wang2019structured,he2019filter,molchanov2019importance,tanaka2020pruning,sun2023simple,frantar2023sparsegpt,yang2025wanda++}.
It not only alleviates computational overhead but also facilitates the adaptation of LLMs to downstream applications with limited labeled data.

Many studies on LLM pruning mainly focus on preserving the generation capability of LLMs \cite{lee2018snip,wang2019structured,kwon2022fast，sun2023simple,frantar2023sparsegpt,fang2023depgraph,ashkboos2024slicegpt,dasu2025attention}.
Early approaches prune the weights or entire substructures in a model that contribute least to its overall performance \cite{mozer1988skeletonization,lecun1989optimal,hassibi1992second,han2015learning},
where they often require architectural modifications or extensive fine-tuning on large datasets to recover any lost performance.
Subsequent studies, such as SparseGPT \cite{frantar2023sparsegpt} and Wanda \cite{sun2023simple}, prune models by measuring the importance of different parts of the model with criteria based on the magnitude of the weights and the computed activation values of the running examples from a calibration dataset.
The calibration data in these studies typically come from general-domain plain text and are used to identify the parameters that allow the model to produce fluent sentences.
These approaches perform well on conventional LLMs such as GPT-3 \cite{brown2020language} and LLaMA \cite{touvron2023llama} that have not undergone instruction-following fine-tuning \cite{ouyang2022training}.
However, by noticing that recent LLMs possess capabilities such as knowledge storage, instruction following, and question answering, and the parameters associated with these functions may not be highlighted by the calibration data,
it is thus of high risk that crucial parameters are potentially wrongly removed and leading to performance degradation across tasks.
\cite{frankle2018lottery,hooker2019compressed}
Particularly, for scenarios such as summarization or domain-specific question answering, model pruning aims not only to shrink the model size but also to retain essential task-specific characteristics.
However, existing pruning approaches that aim at preserving the main ability of text generation for LLMs are not able to guarantee a desirable performance on tasks that require instruction following, since the models may expect various outputs under different instruction contexts with identical input and generic calibration data fails to highlight the weights critical for such conditional behaviors.
Therefore, there is an urgent need for a pruning approach that precisely identifies and preserves task-relevant parameters while effectively compressing the model.

\begin{figure*}[t]
    \centering
    \includegraphics[width=1.0\linewidth, trim=0 10 0 0]{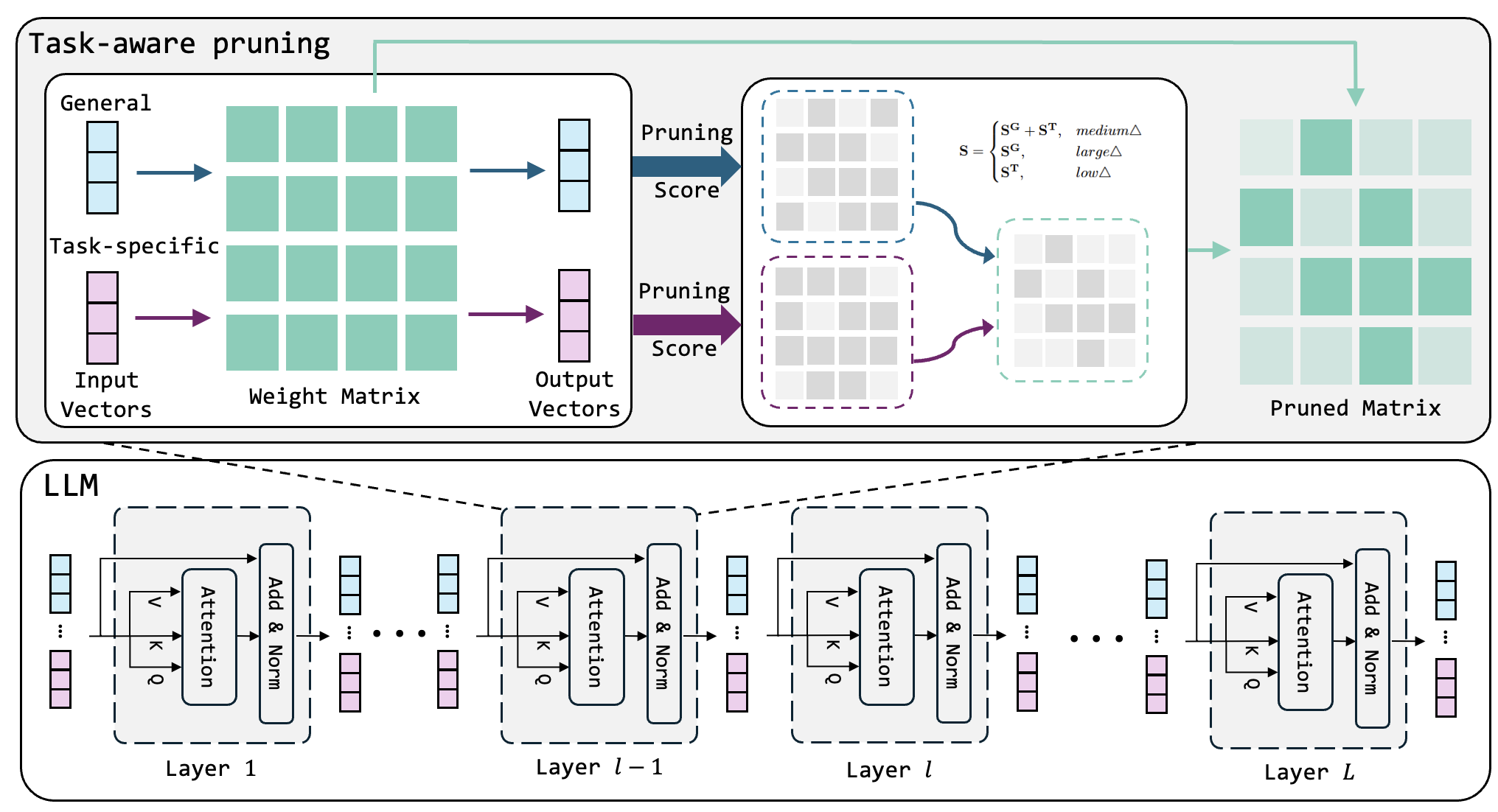}
    \caption{\textcolor{black}{
The overall process of our framework for LLM pruning.
The bottom part presents the LLM.
The top presents the pruning process for a particular parameter matrix in the LLM, where the general and task-specific calibration data (marked by the blue and purple colors, respectively) are used to compute the pruning scores of each parameter.
    }}
\label{fig:model}
\end{figure*}

In this paper, we propose a general LLM pruning framework designed to bridge the gap between generic model compression and task-specific model adaptation. 
Our approach uses two calibration sources (i.e., general and task-specific scores) to compute activation-weighted magnitude scores with a standard pruning backend. 
We then partition channels by activation-norm differences into shared, general-only, and task-only groups, aggregate the corresponding groupwise scores into a mixed importance, and prune the lowest-scoring weights.
Extensive experiments on WikiText-2 and other widely used benchmarks show that our approach effectively removes model parameters while maintaining comparable accuracy and perplexity.

\section{The Approach}

Our approach is a simple yet effective task-aware LLM pruning framework that is operated upon any fundamental pruning algorithms.
%\textcolor{black}{
Specifically, our approach leverages a dual‐source pruning paradigm with calibration datasets from both general‐domain inputs $\mathcal{D}_G$ and task‐specific examples $\mathcal{D}_T$, which together inform parameter importance through direct model activations.
With such data utilization through the LLM, our approach precisely identifies weights whose contributions are not significant across diverse contexts, enabling targeted pruning without compromising the LLM's core capabilities.
Therefore, it is a high‐fidelity pruning strategy that combines general robustness with task‐level precision, ensuring streamlined efficiency at scale.

\subsection{LLM Pruning}
\label{sec:standard-pruning}

In general, pruning 
% sets less relevant parameters to zero while 
controls less relevant parameters in an LLM by
minimizing their impact on downstream task performance, \textcolor{black}{where the distribution of the features for the task data is represented by $\mathcal{D}$.
Assuming we have a calibration dataset $D \sim \mathcal{D}$ with $M$ instances following the feature distribution $\mathcal{D}$, where $D=\{(x,y^*)\}$ with $x$ and $y^*$ denoting the input and the human-annotated output, respectively,
a straightforward solution to measure a model's performance on a downstream task is to compute the loss on the calibration data $D$.
}
Specifically, for a parameter matrix $\mathbf{W}$, the loss function $L(\mathbf{W})$ is defined as the expectation over the \textcolor{black}{feature distribution $\mathcal{D}$}
\begin{equation} \label{eq:loss}
L(\mathbf{W})=\mathbb{E}_{(x,y^*)\sim \mathcal{D}}\bigl[\ell(f_{\mathbf{W}}(x),y^*)\bigr]
\end{equation}
with $\ell(f_{\mathbf{W}}(x),y^*)$ denoting the loss for a single sample between the model's prediction $f_{\mathbf{W}}(x)$ and the ground-truth label $y^*$.
The gradient of the parameter, namely, $\nabla L(\mathbf{W})$, describes the linear response of the loss to small changes in the parameters.
A pruning operation is viewed as applying a perturbation $\Delta \mathbf{W}$ in parameter space.
We define $\Delta W_{ij} = {W}_{ij}^{\mathrm{new}} - {W}_{ij}^{\mathrm{old}}$ to describe such parameter change for $(i,j)$ in the fully connected layer (${W}_{ij}^{\mathrm{new}}$ and ${W}_{ij}^{\mathrm{old}}$ denote the pruned and original weight at position $(i,j)$ in the weight matrix, respectively, and $\Delta W_{ij}$ is the difference of them in the corresponding position of $\Delta \mathbf{W}$).
\textcolor{black}{
When a weight is pruned, it is equivalent to setting its value to zero, which leads to
$\Delta W_{ij} = -W_{ij}$, and all other elements satisfy $\Delta W_{kl} = 0$ for $(k,l)\neq(i,j)$.
}
By performing a multivariate Taylor expansion of the loss at $\mathbf{W}$ and ignoring higher-order terms, one should obtain:
\begin{equation} \label{eq:ori-taylor}
\begin{aligned}
L(\mathbf{W}+\Delta \mathbf{W})
=&\, L(\mathbf{W})
+ \langle\nabla L(\mathbf{W}),\,\Delta \mathbf{W}\rangle
\\
+&\, \tfrac12\,\Delta \mathbf{W}^\top H(\mathbf{W})\,\Delta \mathbf{W}
+ O(\|\Delta \mathbf{W}\|^3)
\end{aligned}
\end{equation}
Therefore, when the original model parameters have been sufficiently trained and are near convergence, $\nabla L(\mathbf{W})\approx 0$, the first-order perturbation is equivalent to be neglected.
The loss difference caused by pruning, $\Delta L(\mathbf{W}) = L(\mathbf{W}+\Delta \mathbf{W}) - L(\mathbf{W})$, approximates to
\begin{equation} \label{eq:ori-second-order}
\Delta L(\mathbf{W})
\approx \tfrac12\,\Delta \mathbf{W}^\top H(\mathbf{W})\,\Delta \mathbf{W}
\end{equation}
where the diagonal elements $H_{ij,ij}$ in the Hessian matrix $H(\mathbf{W}) = \nabla^2 L(\mathbf{W})$ measure the second-order curvature \textcolor{black}{with respect to} the parameter $\mathbf{W}_{ij}$.
Substituting $\Delta W_{ij}=-W_{ij}$ transforms the second-order term into $\tfrac12\,H_{ij,ij}\,\mathbf{W}_{ij}^2$,
which quantifies the 
effect on the loss from pruning weight $W_{ij}$ in terms of second-order sensitivity.
\textcolor{black}{
Therefore, given a pruning rate $r$, the goal of pruning algorithms subjects to find a pruning matrix $\Delta \widehat{\mathbf{W}}$ satisfying the rate $r$ and minimizing the $\Delta L(\mathbf{W})$, which is written as
\begin{equation} \label{eq:opt}
\Delta \widehat{\mathbf{W}} = \arg\min_{\Delta \mathbf{W}}
\tfrac12
\Delta \mathbf{W}^\top H(\mathbf{W}) \Delta \mathbf{W}
\quad
\end{equation}
\text{subject to}
\begin{equation}
\|\Delta \mathbf{W}\|_0 = r \|\mathbf{W}\|_0
\end{equation}
where $\|\cdot\|_0$ counts the number of nonzero entries in a matrix and $\|\Delta \mathbf{W}\|_0 = r \|\mathbf{W}\|_0$ means that exactly an $r$ fraction of the weights are pruned.
% \footnote{\textcolor{black}{We assume all parameters in the original weight $\mathbf{W}$ are non-zero, which is generally the case.}}
%
To facilitate the optimization of Eq. (\ref{eq:opt}), we introduce binary pruning indicators $\alpha_{ij}\in\{0,1\}$ subject to $\sum_{i,j}\alpha_{ij}=r\,\|\mathbf{W}\|_0$ for a particular weight $W_{ij}$, so that 
%$\Delta W_{ij}$ is written as
$\Delta W_{ij}=-\alpha_{ij}\,W_{ij}$.
Therefore, $\tfrac12\,\Delta \mathbf{W}^\top H(\mathbf{W})\,\Delta \mathbf{W}$ is computed by
\begin{equation}
    \begin{aligned}
    \tfrac12\,\Delta \mathbf{W}^\top H(\mathbf{W})\,\Delta \mathbf{W} 
&\approx \tfrac12\sum_{i,j}H_{ij,ij}\,(\Delta W_{ij})^2 \\
&= \tfrac12\sum_{i,j}H_{ij,ij}\,(\alpha_{ij}W_{ij})^2 \\
&= \tfrac12\sum_{i,j}H_{ij,ij}\,W_{ij}^2\,\alpha_{ij}
\end{aligned}
\end{equation}
where off-diagonal Hessian entries are ignored.
Thus, the objective Eq. (\ref{eq:opt}) is rewritten as finding a set of optimal $\{\alpha_{ij}^*\}$ that satisfies the pruning rate, which is formulated as
\begin{equation} \label{eq:alpha-goal}
\{\alpha_{ij}^*\}
=\arg\min_{\{\alpha_{ij}\}}
\sum_{i,j}H_{ij,ij}\,W_{ij}^2\,\alpha_{ij}
\end{equation}
subject to
\begin{equation}
    \sum_{i,j}\alpha_{ij}=r\,\|\mathbf{W}\|_0
\end{equation}
As a result, the value of $H_{ij,ij}\,W_{ij}^2$ determines whether the weight $W_{ij}$ should be pruned or not.
We use the term $S_{ij}$ to represent $H_{ij,ij}\,W_{ij}^2$ by 
\begin{equation} \label{eq:weight}
S_{ij} \;:=\; H_{ij,ij}\,W_{ij}^2
\end{equation}
and name $S_{ij}$ as the important score to measure the effect of the parameter $W_{ij}$ on the model's performance, where a higher score means that the parameter has a more powerful effect on the loss.
Therefore, the optimal strategy is to prune the weights whose importance scores are smallest, until exactly an $r$ fraction of the parameters is selected.
Equivalently, one is able to set 
\begin{equation} \label{eq:alpha}
\alpha_{ij}^* = 
\begin{cases}
1, & S_{ij} \le \tau \\
0, & S_{ij} > \tau
\end{cases}
\quad
\#\{(i,j)\mid s_{ij}\le\tau\} = r\,\lvert\mathbf{W}\rvert_0
\end{equation}
where $\#\{\cdot\}$ denotes the number of elements in a set, so that $\Delta\widehat{\mathbf{W}}$ is fully determined by thresholding the scores $s_{ij}$.
}

\textcolor{black}{
To illustrate the pruning process, we utilize a representative algorithm named Wanda \cite{sun2023simple} as an example.
Specifically, for a standard fully connected layer, Wanda computes $H_{ij,ij}$ by
\begin{equation} 
H_{ij,ij}
\approx\tfrac1M\sum_{m=1}^M\bigl(x_j^{(m)}\bigr)^2
\end{equation}
where the activation vector for channel $j$ over the input of the calibration set $\{x^{(m)}\}_{m=1}^M$ is denoted by $\mathbf{X}_j=[x_j^{(1)},\dots,x_j^{(M)}]$.
Therefore, the important score $S_{ij}$ is computed by
\begin{equation} \label{eq:wanda-base}
S_{ij}=\tfrac1M\sum_{m=1}^M\bigl(x_j^{(m)}\bigr)^2 \cdot W_{i,j}^2
\end{equation}
where $\sum_{m=1}^M\bigl(x_j^{(m)}\bigr)^2$ is the square of the activation L2 norm $\|\mathbf{X}_j\|_2=\sqrt{\sum_{m=1}^M(x_j^{(m)})^2}$, reflecting the average activation strength of channel $j$ and 
${W}_{ij}^2$ is the square of the weight of the connection $(i,j)$, representing its magnitude.
This score $S_{ij}$ combines weight magnitude with activation strength, with a higher value indicating a more important role the weight plays in the LLM. 
In other words, given a pruning rate $r$, the parameters whose score $S_{ij}$ is ranked on the bottom-$r$ are pruned.
}

\subsection{Task-Specific Pruning}

For task-specific LLM utilization, the most obvious characteristic is that
\textcolor{black}{feature} distribution differs from that of general-purpose scenarios.
The pruning thus must account for task-related features to ensure that calibration-driven pruning aligns with such distribution and task requirements.
\textcolor{black}{
Therefore, given feature distribution from the general domain $\mathcal{D}_G$ and the task-specific domain $\mathcal{D}_T$, the goal is to utilize two calibration datasets with one (denoted as $D_G$) from the general domain (i.e., $D_G \sim \mathcal{D}_G$) and the other (denoted as $D_T$) from the target task domain (i.e., $D_T \sim \mathcal{D}_T$) for pruning.
Therefore, the loss function of Eq. (\ref{eq:loss}) is written as
\begin{equation} \label{eq:task-loss}
L(\mathbf{W})=\mathbb{E}_{(x,y^*)\sim \mathcal{D}_{G} \cup \mathcal{D}_{T} }\bigl[\ell(f_{\mathbf{W}}(x),y^*)\bigr]
\end{equation}
Since $\mathcal{D}_{G}$ and $\mathcal{D}_{T}$ may not be independent with each other, 
we define three disjoint subsets of the calibration data, namely, $\mathcal{D}_{GT}$, $\mathcal{D}_{G-}$, and $\mathcal{D}_{T-}$ by
\begin{equation}
\begin{aligned}
\mathcal{D}_{GT} = \mathcal{D}_{G} \cap \mathcal{D}_{T} \\
\mathcal{D}_{G-} = \mathcal{D}_{G} \setminus \mathcal{D}_{GT} \\
\mathcal{D}_{T-} = \mathcal{D}_{T} \setminus \mathcal{D}_{GT}
\end{aligned}
\end{equation}
where $\mathcal{D}_{GT}$ is the subset of samples shared by both the general domain and the task-specific domain, $\mathcal{D}_{G-}$ is the subset of samples unique to the general domain, and $\mathcal{D}_{T-}$ is the subset of samples unique to the task-specific domain.
Therfore, $\mathcal{D}_{GT}$, $\mathcal{D}_{G-}$, and $\mathcal{D}_{T-}$ are mutually disjoint and we decompose Eq. (\ref{eq:task-loss}) to
\begin{equation}
    L(\mathbf{W}) \;=\; L_{GT}(\mathbf{W}) \;+\; L_{G-}(\mathbf{W}) \;+\; L_{T-}(\mathbf{W})
\end{equation}
where $L_{GT}(\mathbf{W})$, $L_{G-}(\mathbf{W})$, and $L_{T-}(\mathbf{W})$ are the losses for the shared, general-only and the task-specific-only domains and are computed by
\begin{equation}
\begin{aligned}
L_{GT}(\mathbf{W}) &= \mathbb{E}_{(x,y^*) \sim \mathcal{D}_{GT}}\bigl[\ell(f_{\mathbf{W}}(x),y^*)\bigr]\\
L_{G-}(\mathbf{W}) &= \mathbb{E}_{(x,y^*) \sim \mathcal{D}_{G-}}\bigl[\ell(f_{\mathbf{W}}(x),y^*)\bigr]\\
L_{T-}(\mathbf{W}) &= \mathbb{E}_{(x,y^*) \sim \mathcal{D}_{T-}}\bigl[\ell(f_{\mathbf{W}}(x),y^*)\bigr]
\end{aligned}
\end{equation}
Following Eq. (\ref{eq:ori-taylor}), we expand the mixed loss at $\mathbf{W}$ by a second-order Taylor series and neglect the first-order term, and obtain 
\begin{equation} \label{eq:final prune}
\begin{aligned}
    \Delta L(\mathbf{W})
\approx& \Delta \mathbf{W}^\top
H_{GT} \Delta \mathbf{W} + \Delta \mathbf{W}^\top H_{G-} \Delta \mathbf{W} \\ +& \Delta \mathbf{W}^\top H_{T-} \Delta \mathbf{W}
\end{aligned}
\end{equation}
where 
\begin{equation}
\begin{aligned}
    H_{GT} = \nabla^2 L_{GT}(\mathbf{W}) \\ 
    H_{G-} = \nabla^2 L_{G-}(\mathbf{W}) \\ 
    H_{T-} = \nabla^2 L_{T-}(\mathbf{W})
\end{aligned}
\end{equation}
are the Hessian matrices for the shared, general-only, and the task-specific-only domains, respectively.
Following the process of Eq. (\ref{eq:weight})-(\ref{eq:alpha}), task-specific pruning computes the important score $S_{ij}$ by the sum of the important scores from the shared, general-only, and the task-specific-only domains (which are denoted as $S_{ij}^{GT}$, $S_{ij}^{G-}$ and $S_{ij}^{T-}$, respectively) by
\begin{align} \label{eq:mix}
    S_{ij} 
    &= S_{ij}^{GT} + S_{ij}^{G-} + S_{ij}^{T-}
\end{align}
To compute the pruning scores for different groups, we need to divide the data features from the general domain and the task-specific domains into the three groups, i.e., the features shared by both the general and task-specific domains, the ones unique to the general domain, and the ones unique to the task-specific domain.
}

\textcolor{black}{
To this end, we design an approach based on the difference in activation magnitudes.
Specifically, for channel $j$ in a given layer (assume the total number of channels is $d$), we compute its activation norms over the general domain calibration set, $|x_j^{(G)}|_2$, and over the task-specific domain calibration set, $|x_j^{(T)}|_2$, and then form the difference score $\Delta_j$ by
\begin{equation}
   \Delta_j = |x_j^{(G)}|_2 - |x_j^{(T)}|_2
\end{equation}
We then sort all channels by $\Delta_j$ in descending order and divide them according to a positive hyperparameter $\alpha$: channels with $\Delta_j > \alpha$ are classified as general-only, those with $- \alpha < \Delta_j < \alpha$ are task-only, and those with $\Delta_j$ between $-\alpha$ and $\alpha$ are treated as shared.
Intuitively, this means that if a channel’s activation in the general domain exceeds that in the task-specific domain by more than $\alpha$, it is general-only; if it is lower by more than $\alpha$, it is task-only; and if the difference is within the range $[-\alpha, \alpha]$, it is classified as shared.
For each group, we use the following approach to compute the pruning scores for the parameters.
First, for the general-domain-only channels (index set $\mathcal{J}_{G-}$), we compute the pruning score using the examples from the general domain.
Similarly, for the task-specific domain channels (index set $\mathcal{J}_{T-}$), we compute the pruning score using the examples from the task-specific domain.
For shared channels (index set $\mathcal{J}_{GT}$), we compute the pruning score using the sum of pruning scores from both the general and the task-specific domain.
Therefore, $S_{ij}^{GT}$, $S_{ij}^{G-}$, and $S_{ij}^{T-}$ are computed by
\begin{equation}
\begin{aligned}
    S_{ij}^{GT} &= s_{ij}^G + s_{ij}^T, \quad j \in \mathcal{J}_{GT} \\
    S_{ij}^{G-} &= s_{ij}^G, \quad j \in \mathcal{J}_{G-} \\
    S_{ij}^{T-} &= s_{ij}^T, \quad j \in \mathcal{J}_{T-}
\end{aligned}
\end{equation}
where $s_{ij}^G$ and $s_{ij}^T$ are the pruning scores obtained through the standard pruning process using the general domain and the task-specific domain data, respectively.
Because $S_{i,j}^{GT} = S_{i,j}^{G} + S_{i,j}^{T}$ holds for shared channels, this naturally makes the shared score twice that of any exclusive score, reflecting its higher importance and matching the intuition that the parameters for the shared part of the domains should be more important than the parameters for only one domain.
Finally, we sort all mixed importance scores $S_{i,j}$ in the entire model from smallest to largest and prune away the lowest $r$ fraction of parameters according to the target sparsity rate $r$.
}

\textcolor{black}{
Using the Wanda pruning algorithm as an example.
For each parameter $W_{ij}$, Wanda computes a pruning score $S_{ij}$ following Eq. (\ref{eq:wanda-base}).
This score corresponds to the effect of pruning the weight $W_{ij}$ on the loss under the task-specific settings, balancing the parameter sensitivities of the general and task-specific domains.
We sort all parameters based on their mixed pruning score $S_{ij}$.
We iteratively prune the weights with the lowest scores until the predetermined pruning ratio $r$ is met.
}

\begin{table*}[t]
\centering
\caption{
\textcolor{black}{
Overall results on WikiText-2 (WT-2) (PPL), MMLU, MedQA, and ARC for Qwen-3 (32B) at various compression ratios (CR), where the performance of the original LLM without pruning is also reported for reference.
Lower perplexity on WT-2 and higher accuracy on other datasets indicate better performance, which are marked by $\downarrow$ and $\uparrow$, respectively.
``Avg.'' refers to the average performance on MMLU, MedQA, and ARC.
}
}
\label{tab:overall}
\begin{tabular}{lcl||c||ccc|c}
\toprule
\textbf{Model} & \textbf{CR} & \textbf{Approach} & \textbf{WT-2} $\downarrow$
& \textbf{MMLU} $\uparrow$
& \textbf{MedQA} $\uparrow$
& \textbf{ARC} $\uparrow$
& \textbf{Avg.} $\uparrow$ \\
\midrule
\multirow{8}{*}{\makecell[c]{Qwen-3\\(32B)}} &
- & - 
& 6.45 
& 80.77 
& 74.55 
& 62.07 
& 72.46 \\
\cmidrule{2-8}
& \multirow{2}{*}{50\%} 
& Wanda 
& 8.38 
& \textbf{76.93} 
& 69.05 
& 56.06 
& 67.34 \\
& & Ours 
& 8.45 
& 76.58 
& \textbf{70.15} 
& \textbf{58.36} 
& \textbf{68.36} \\
\cmidrule{2-8}
& \multirow{2}{*}{75\%} 
% & 
& Wanda 
& 23.72 
& 41.56 
& 31.34 
& 29.10 
& 34.00 \\
& & Ours 
& 28.47 
& \textbf{43.13} 
& \textbf{33.46} 
& \textbf{29.52}
& \textbf{35.37} \\
\cmidrule{2-8}
& \multirow{2}{*}{90\%} 
& Wanda 
& 1856 
& 24.78 
& 25.92 
% & 27.78 
% & 20.05 
& 23.71 
& 24.80 \\
& & Ours 
& 1281 
& \textbf{25.34} 
& \textbf{27.33} 
& \textbf{23.98} 
& \textbf{25.55} \\
\bottomrule
\end{tabular}
\end{table*}

\section{Experiment Settings}

\subsection{Datasets}

To comprehensively evaluate our pruning approach, we conduct experiments on several widely used benchmark datasets that target different aspects of model performance.
For assessing model fluency and general language generation capabilities, we use the WikiText-2 dataset \cite{merity2016pointer}\footnote{\url{https://huggingface.co/datasets/mindchain/wikitext2}},
which is a widely used corpus that contains high-quality articles from Wikipedia, making it suitable for evaluating the naturalness and coherence of LLM outputs.
We also employ a set of zero/few-shot evaluation datasets to measure the LLMs’ generalization ability across various tasks.
These datasets\footnote{Specifically, we obtain \textcolor{black}{MMLU, MedQA, and ARC} from \url{https://huggingface.co/datasets/cais/mmlu}, \url{https://huggingface.co/datasets/GBaker/MedQA-USMLE-4-options}, 
and \url{https://huggingface.co/datasets/allenai/ai2_arc}, 
respectively.} include:
\begin{itemize}[leftmargin=1em]
\item \textbf{MMLU} \cite{hendrycks2020measuring}: A multiple-choice benchmark covering 57 subjects (e.g., humanities, STEM, professional topics), testing both factual knowledge and reasoning ability.
\item \textbf{MedQA} \cite{jin2021disease}: A medical question-answering (QA) dataset derived from United States Medical Licensing Examination (USMLE) questions in four-choice format, covering clinical diagnosis and treatment scenarios in the medical domain.
\item \textbf{ARC} \cite{clark2018think}: The AI2 Reasoning Challenge dataset with grade-school science questions in multiple-choice format, requiring qualitative reasoning across physics, biology, and chemistry.
\end{itemize}
We utilize the standard split of these datasets for evaluating the performance of different LLMs.

For the \textcolor{black}{calibration} datasets, we utilize the English C4 \cite{raffel2020exploring} as the general domain data\footnote{\url{https://huggingface.co/datasets/c4}} (i.e., $D_G$),
which is a cleaned, deduplicated version of the Common Crawl web corpus.
We also utilize the training splits of the MMLU, MedQA, and ARC for task-aware \textcolor{black}{calibration} dataset (i.e., $D_T$), which ensures that the learned importance scores are reflective of \textcolor{black}{the task-specific feature} distribution.
For C4, we randomly sample 128 examples; for the task-aware \textcolor{black}{calibration} dataset, we also randomly sample 128 examples from the union of them to maintain a balanced evaluation setup.

\subsection{Fundamental Pruning Algorithms}

The fundamental pruning algorithm used in our experiment is Wanda \cite{sun2023simple}.
Wanda employs a pruning metric by multiplying weight magnitude with input activation norms to directly assess connection importance without requiring any fine-tuning.
The algorithm supports structured and unstructured pruning.
Our approach is validated on the algorithm to demonstrate the generality and effectiveness of task-aware pruning across different paradigms.

\subsection{Implementation Details}

\textcolor{black}{
We run experiments with state-of-the-art LLMs, i.e., the Qwen-3 \cite{bai2023qwen} series\footnote{\textcolor{black}{The LLM is obtained from the official website on HuggingFace \url{https://huggingface.co/}.}}. 
In our experiments, we adopt the Qwen-3 (32B) version of Qwen-3.
Specifically, the Qwen-3 (32B) version has 64 layers of Transformer with a 5120-dimensional hidden size.
We run experiments with both unstructured and structured pruning to validate the robustness of our task-aware approach.
For unstructured pruning, we explore three compression ratios, $r = 50\%, 75\%,\text{ and } 90\%$, where $r$ denotes the fraction of parameters pruned (e.g., $r = 0.75$ prunes 75\% of the weights).
For structured pruning, we employ 2:4 and 4:8 N:M sparsity patterns, which enforce that within each contiguous block of 4 (resp.\ 8) parameters, exactly 2 (resp.\ 4) parameters are retained, thereby achieving 50\% sparsity in a hardware-friendly layout.
For the weight $\alpha$ to determine whether a channel's activation belongs to the shared or a particular domain, we  try different values (i.e., 0.01, 0.05, 0.1, 0.2, 0.5, 1.0, and 2.0) in the analysis.
}
For evaluation, we utilize perplexity (PPL) \cite{jelinek1977perplexity} to measure language fluency on the WikiText-2 dataset, and evaluate by accuracy\footnote{\url{https://github.com/EleutherAI/lm-evaluation-harness}.} under the zero-shot setting on MMLU, MedQA, and ARC.

\section{Results and Analysis}

%%%%%%%%%%%%%%%%%%%%%%%%%%%%%%%%%%%%%%%%%%%%%%%%%%%%%%%%%%%%%%%%%%%%%%%%%%%%%%
\subsection{Overall Results}
%%%%%%%%%%%%%%%%%%%%%%%%%%%%%%%%%%%%%%%%%%%%%%%%%%%%%%%%%%%%%%%%%%%%%%%%%%%%%%

\textcolor{black}{
We summarize the overall performance of our task-aware pruning approach and the baseline Wanda across multiple compression ratios in Table \ref{tab:overall}.
Several key observations are drawn from these results.
First, on average, our approach consistently achieves better performance than Wanda under all pruning ratios.
At moderate sparsity levels such as 50\% and 75\%, our approach shows clear improvements on most downstream tasks (including MMLU, MedQA, and ARC), while maintaining comparable perplexity on WikiText-2, indicating that the integration of task-specific information effectively enhances model capability without harming general fluency.
Second, as the pruning ratio reaches 90\%, the performance of all models naturally declines, yet our approach consistently yields higher average accuracy than Wanda across most settings, demonstrating its robustness under aggressive compression.
Overall, these results confirm that our task-aware pruning strategy achieves a better balance between task performance than the baseline model, making it a reliable solution for compressing LLMs without severely degrading their downstream capability.
}

\begin{table*}[t]
\centering
\caption{
\textcolor{black}{
The results of the Qwen 3 (32B) pruned by the original Wanda and our approach with Wanda under different pruning configurations.
``50\%'' means the unstructured pruning where 50\% parameters are pruned.
``2:4'' and ``4:8'' are structured pruning settings which also lead to 50\% pruned parameters.
}
}
\label{tab:usage}
\begin{tabular}{lc|c||ccc|c}
\toprule
\textbf{PS} & \textbf{Approach} & \textbf{WT-2} $\downarrow$
& \textbf{MMLU} $\uparrow$
& \textbf{MedQA} $\uparrow$
& \textbf{ARC} $\uparrow$
& \textbf{Avg.} $\uparrow$ \\
\midrule
\multirow{2}{*}{50\%} 
& Wanda
& 8.38 
& \textbf{76.93} 
& 69.05 
& 56.06 
& 67.34 \\
& Ours 
& 8.45 
& 76.58 
& \textbf{70.15} 
& \textbf{58.36} 
& \textbf{68.36} \\
\midrule
\midrule
\multirow{2}{*}{2:4} & Wanda 
& 8.26 & 68.78 & 58.94 & 52.21 & 59.97 \\
& Ours 
& 8.53 & \textbf{69.13} & \textbf{59.23} & \textbf{52.56} & \textbf{60.30} \\
\midrule
\multirow{2}{*}{4:8} & Wanda 
& 7.67 & 73.27 & 65.36 & 54.23 & 64.28 \\
& Ours 
& 7.82 & \textbf{73.62} & \textbf{65.67} & \textbf{54.69} & \textbf{64.66} \\
\bottomrule
\end{tabular}
\end{table*}

%%%%%%%%%%%%%%%%%%%%%%%%%%%%%%%%%%%%%%%%%%%%%%%%%%%%%%%%%%%%%%%%%%%%%%%%%%%%%%
\subsection{The Performance on Structured Pruning}
%%%%%%%%%%%%%%%%%%%%%%%%%%%%%%%%%%%%%%%%%%%%%%%%%%%%%%%%%%%%%%%%%%%%%%%%%%%%%%

\textcolor{black}{
To further examine the robustness of our task-aware pruning approach, we conduct additional experiments on structured pruning configurations.
Specifically, we test two hardware-friendly sparsity patterns—2:4 and 4:8 N:M pruning—each achieving approximately 50\% sparsity.
For comparison, we also include the 50\% unstructured pruning results in Table \ref{tab:usage}.
This allows a fair evaluation of our approach under both flexible and constrained pruning regimes.
Several observations can be made from these results.
First, our approach consistently outperforms the original Wanda across all pruning configurations, including both structured and unstructured settings.
This confirms that introducing task-aware calibration effectively improves model retention of important parameters regardless of the pruning form.
Second, the performance gain is steady across datasets such as MMLU, MedQA, and ARC, indicating that the proposed method generalizes well across different evaluation domains and compression types.
Third, while structured pruning generally leads to slightly lower absolute performance than unstructured pruning (owing to its inherent constraints on parameter selection) our approach still delivers consistent advantages over Wanda.
This demonstrates that our method remains effective even when pruning flexibility is limited by hardware-aware sparsity patterns.
Overall, these results validate that our task-aware pruning framework extends naturally to structured pruning scenarios and continues to enhance performance beyond existing baseline techniques.
}

\begin{figure}
\centering
\includegraphics[width=1\linewidth]{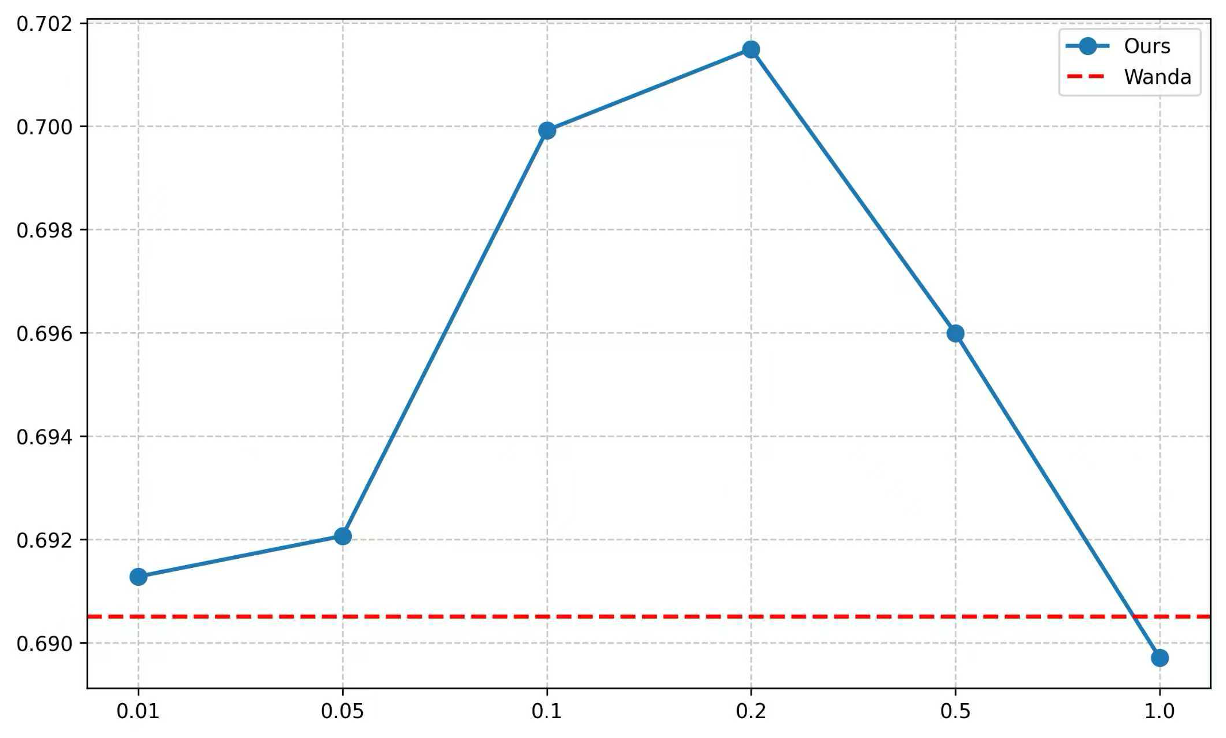}
\caption{
\textcolor{black}{
Performance of the pruned Qwen-3 (32B) on MedQA at 50\% sparsity under different $\alpha$ values (blue curve).
The dashed red line denotes the Wanda baseline.
The parameter $\alpha$ controls how strictly the model separates shared and task-specific channels during pruning.
}
}
\label{fig:score}
\end{figure}

%%%%%%%%%%%%%%%%%%%%%%%%%%%%%%%%%%%%%%%%%%%%%%%%%%%%%%%%%%%%%%%%%%%%%%%%%%%%%%
\subsection{Effect of the Threshold for Pruning}
% %%%%%%%%%%%%%%%%%%%%%%%%%%%%%%%%%%%%%%%%%%%%%%%%%%%%%%%%%%%%%%%%%%%%%%%%%%%%%%

\textcolor{black}{
We investigate how the threshold $\alpha$ influences model performance, using MedQA as a representative downstream task and 128 calibration examples for pruning.
The parameter $\alpha$ determines how feature channels are partitioned, where smaller $\alpha$ values make the boundary between shared and task-specific channels tighter, while larger $\alpha$ values broaden the shared region and reduce the distinctiveness of task-specific components.
Figure \ref{fig:score} reports the accuracy of our approach on MedQA under different $\alpha$ values for a 50\% pruning rate (the blue curve), with the Wanda baseline shown as a dashed line.
When $\alpha$ is very small ($\alpha\!\le\!0.05$), the model barely distinguishes shared from task-specific parameters, causing the pruning to behave similarly to the generic Wanda baseline, and the improvement is marginal.
As $\alpha$ increases, the separation becomes clearer, allowing the model to better balance shared linguistic knowledge and task-specific representations.
This leads to a noticeable performance gain that peaks around $\alpha\!=\!0.2$.
However, when $\alpha$ continues to grow ($\alpha\!\ge\!0.5$), more parameters tend to be classified as shared, which weakens the model’s ability to capture task-relevant specialization and causes performance degradation.
An intermediate value around $\alpha\!=\!0.2$ provides the most effective balance, enabling our approach to maintain broad knowledge coverage while preserving task-specific discriminability during pruning.
}

\begin{figure}
\centering
\includegraphics[width=1\linewidth]{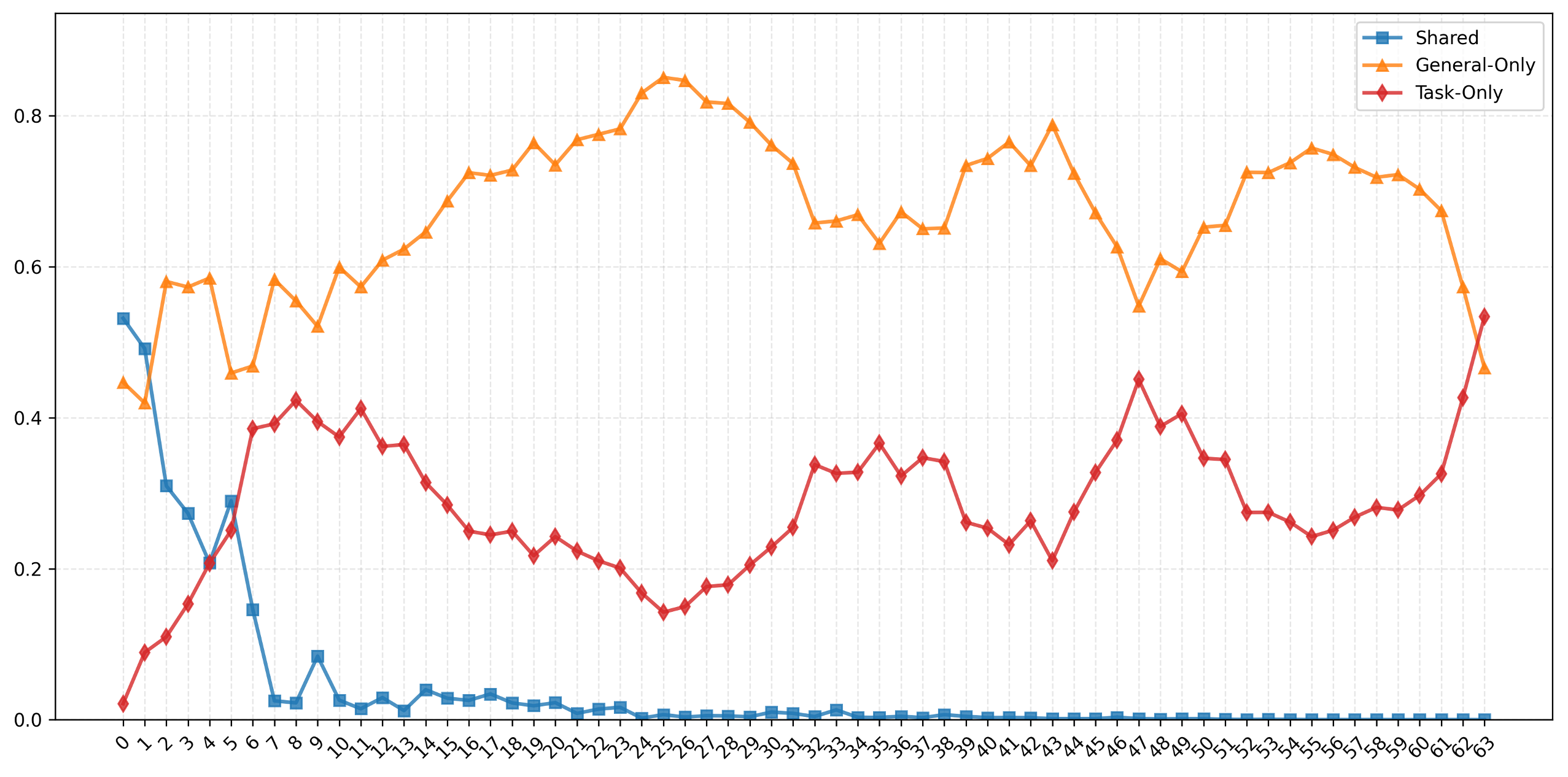}
\caption{
\textcolor{black}{
Distribution of parameter groups across Transformer layers in Qwen-3 (32B).
The x-axis represents the layer index, and the y-axis shows the proportion of parameters categorized as shared (blue), general-only (orange), and task-only (red).
}
}
\label{fig:score}
\end{figure}

%%%%%%%%%%%%%%%%%%%%%%%%%%%%%%%%%%%%%%%%%%%%%%%%%%%%%%%%%%%%%%%%%%%%%%%%%%%%%%
\subsection{Visualization of Parameter Type Distribution during Pruning}
%%%%%%%%%%%%%%%%%%%%%%%%%%%%%%%%%%%%%%%%%%%%%%%%%%%%%%%%%%%%%%%%%%%%%%%%%%%%%%

\textcolor{black}{
To better understand how task-aware pruning differentiates parameter roles across the model depth, we visualize the proportion of parameters classified as shared, general-only, and task-only in each Transformer layer of Qwen-3 (32B), as shown in Figure \ref{fig:score}.
The x-axis denotes the layer index, and the y-axis represents the fraction of parameters assigned to each group.
Several trends are observed.
First, in the lower layers, the majority of parameters fall into the shared category, while the proportions of general-only and task-only parameters remain small.
This suggests that the lower layers mainly capture fundamental linguistic and syntactic patterns that are common to both domains, with limited domain differentiation.
Second, as the layer depth increases, the share of general-only and task-only parameters rises sharply, and their divergence begins to dominate the model’s representation space.
The general-only parameters (orange curve) maintain a strong presence across middle layers, indicating that general linguistic knowledge continues to play a central role in intermediate representations.
Meanwhile, task-only parameters (red curve) gradually increase in deeper layers, showing that high-level domain-specific reasoning and adaptation are formed near the top of the model.
Finally, the shared proportion (blue curve) decreases toward upper layers and becomes negligible after a certain depth, reflecting that high-level modules specialize either toward task-specific behavior or general-domain understanding.
Overall, this layer-wise pattern supports our intuition that the proposed task-aware grouping mechanism is able to identify the hierarchical specialization structure of the LLM: lower layers encode shared generality, middle layers balance both domains, and upper layers focus on task-specific adaptation.
}

\section{Related Work}

\textcolor{black}{
Pruning aims to reduce the size of a model while preserving its capabilities to perform various tasks.
The pruning task has attracted attention from existing studies for decades \cite{mozer1988skeletonization,lecun1989optimal,hassibi1992second,han2015learning,louizos2017learning,wang2019structured,ma2023llm,nova2023gradient,wu2024iterative}.
In the era of LLM, where high-quality text representations are crucial for downstream performance \cite{mikolov2013efficient,ijcai2018-607,peters-etal-2018-deep,devlin2019bert,brown2020language,song2021zen,achiam2023gpt,touvron2023llama,bai2023qwen} and model scales have expanded dramatically, pruning becomes increasingly important for mitigating the computational and memory costs of training and inference.
Conventional studies in neural network pruning utilize straightforward magnitude-based approaches, which rank weights by absolute value and prune the smallest ones, demonstrating that a significant fraction of parameters can be discarded with minimal loss in accuracy \cite{han2015learning,han2015deep,hu2016network,kuzmin2023pruning,an2024fluctuation}. 
There are studies that utilize a dynamic sparsity schedule mechanism that allows weights to be masked and unmasked during fine-tuning, improving over static magnitude thresholds \cite{gale2019state,evci2020rigging,sanh2020movement,sokar2021dynamic,hoefler2021sparsity}. 
These pruning approaches mainly focus on the magnitude of the weights, where less attention is paid to the running data that the models are expected to be applied to.
Therefore, there are studies (e.g., SparseGPT \cite{frantar2023sparsegpt} and Wanda \cite{sun2023simple}) that leverage weight magnitude and activation values from running data from a calibration set to measure the weight importance so as to improve pruning.
To further advance these approaches, subsequent research introduces enhancements in activation-aware magnitude metrics, regional gradient-based importance scoring, and iterative second-order recovery \cite{frantar2023sparsegpt,zimmer2023perp,yang2025wanda++,wu2024iterative,bai2024sparsellm}.
These approaches primarily focus on preserving fluent natural language generation while placing comparatively less emphasis on performance in specific tasks.
Compared to previous studies, our approach focuses on task-aware LLM pruning, which utilizes both general-domain and task-specific data to better identify the unimportant parameters to be pruned.
}

\section{Conclusion}

\textcolor{black}{
In this paper, we propose a frustratingly easy task-aware pruning approach for LLMs.
Specifically, we use two calibration datasets from the general and task-specific domains and apply them within existing pruning algorithms (e.g., Wanda) to compute activation-weighted magnitude scores for each parameter.
We then partition channels by activation-norm differences into shared, general-only, and task-only groups through a simple threshold, and combine their respective importance scores into a mixed score that guides pruning.
We evaluate our approach using the widely adopted Wanda algorithm.
Experimental results confirm that our task-aware pruning effectively preserves domain- and task-specific capabilities, consistently outperforming standard pruning baselines across multiple benchmarks.
}

% In the unusual situation where you want a paper to appear in the
% references without citing it in the main text, use \nocite
\nocite{langley00}

\ifCLASSOPTIONcaptionsoff
  \newpage
\fi

\bibliographystyle{IEEEtran}
\bibliography{reference,purning}

\newpage

\appendices

\end{document}